\documentclass[letterpaper, 10 pt, conference]{ieeeconf}  

\IEEEoverridecommandlockouts                              

\usepackage{graphicx} 
\usepackage{amsmath} 
\usepackage{amssymb}  
\usepackage{cite}
\usepackage{booktabs}
\usepackage{multirow}
\usepackage{xcolor}
\usepackage{makecell}
\usepackage[caption=false,font=footnotesize]{subfig}

\newtheorem{definition}{\textbf{Definition}}

\newtheorem{remark}{\rm\textbf{Remark}}
\newtheorem{proposition}{\rm\textbf{Proposition}}

\usepackage{hyperref}

\hypersetup{%
  colorlinks=true,%
  linkcolor={black},
  citecolor={black},
  urlcolor={black}
  bookmarksnumbered=true,%
  bookmarksopen=true}

\title{\LARGE \bf
Safe Polytope-in-Polytope Motion Planning and Control with Control Barrier Functions
}

\author{Alejandro Gonzalez-Garcia$^\dagger$, Dries Dirckx$^\dagger$, Jan Swevers and Wilm Decr\'e
\thanks{This work was supported by the Flanders Make SBO projects ARENA (Agile \& Reliable Navigation) and LearnOpTra (Learning
meets optimization for robust and multimodal trajectory planning).}
\thanks{All authors are with MECO Research Team, Department of Mechanical Engineering, KU Leuven, Belgium and Flanders Make@KU Leuven, Belgium.
         {\tt\small \{alex.gonzalezgarcia, dries.dirckx, jan.swevers, wilm.decre\}@kuleuven.be}}%
\thanks{$^\dagger$A. Gonzalez-Garcia and D. Dirckx are equal contributors to this work.}         
}

\begin{document}

\maketitle
\thispagestyle{empty}
\pagestyle{empty}

\begin{abstract}
Autonomous mobile robots operating in tight environments require motion planning frameworks that account for the physical footprint of the robot. Simplifying the geometry to a point or a circle is conservative and discards information needed to successfully and safely traverse narrow passages. This work proposes a safe local motion planning and control method that guarantees that a polytopic robot footprint stays inside a continuously updated convex free-space region. The containment condition is formulated as a set of discrete-time control barrier function constraints within a model predictive controller. The number of safety constraints depends on the complexity of the local free-space geometry and the robot shape, instead of the number of obstacles. The proposed free-space formulation does not need any obstacle detection or segmentation. A comparative analysis against a polytope-based obstacle avoidance formulation confirms favorable scaling up to a reduction of 91$\times$ in computation time as the number of obstacles increases. The approach is validated in simulation with an autonomous surface vehicle and on hardware with a non-holonomic mobile robot, using both occupancy grids and LiDAR sensing. The experiments demonstrate safe real-time motion planning and control at 10~Hz on an onboard embedded computer, including reactive avoidance of dynamic obstacles.
\end{abstract}





\section{Introduction}
\label{sec:introduction}

Autonomous mobile robots operate in diverse and complex environments, requiring frameworks capable of guaranteeing safe motion planning and control, especially within constrained spaces. Narrow passages are present in a multitude of applications, from indoor environments with limited clearance, or industrial facilities with dense layouts, to inland waterways with irregular canal walls. In such contexts, accurate consideration of both the physical footprint of the robot and the enclosing environment is an essential element to maneuver safely and successfully. Methods that simplify the robot's geometry to points or circles become overly conservative, discarding vital information to navigate through tight spaces. This creates a need for frameworks that consider the actual geometry of both the robot and the surrounding free space.\newline

Classic motion planning methods rely on graph search through occupancy grid maps~\cite{occupancygrid}. Such graph-based methods, including \textit{A}*~\cite{astar} and Dijkstra's algorithm~\cite{dijkstra1959note}, use heuristics to compute collision-free global paths. Modern extensions involve motion primitives to consider kinodynamic feasibility, or incorporate the robot footprint into the heuristic~\cite{macenski2020marathon2}. However, these methods result in discontinuous or non-smooth solutions. Sampling-based methods, built on probabilistic roadmaps~\cite{508439} or rapidly-exploring random trees~\cite{11282962}, can produce kinodynamically feasible trajectories, but may lack safety guarantees during execution and become intractable in cluttered or constrained spaces. Hence, there is a need for local methods to follow these approximate paths and guarantee online safety. Traditional local methods, such as approaches based on potential fields~\cite{Khatib1985} and velocity obstacles~\cite{Fiorini1993}, are reactive and known to easily get trapped in local minima or deadlocks, which hinders deployment in real-world settings.\\

Model predictive control (MPC)~\cite{MAYNE2000789} integrates dynamics, constraints, and objectives into an optimal control problem (OCP), making it a suitable tool for real-time robot navigation~\cite{panoc2018ecc,9385847,11419776}. However, standard distance-based constraint formulations in MPC require long prediction horizons to produce effective obstacle avoidance behaviors. Control barrier functions (CBFs) address this by encoding safety through set invariance~\cite{ames2019cbftheory,xiao2021high}, and have been widely applied for obstacle avoidance~\cite{liu2025acc, conecbf2024acc}. By incorporating discrete-time CBF constraints into MPC~\cite{Agrawal2017,dcbf2021}, safety can be maintained with shorter horizons, yielding a tractable strategy for real-time, safe control.\newline

While MPC-CBF formulations ensure safety effectively, they commonly assume point-mass or circular robot footprints. For non-circular robots, accounting for the true geometry within an OCP requires either simplifying the shape or introducing significant constraint complexity. Approaches such as~\cite{10886807} propose ellipsoidal representations to reduce conservatism, but remain limited in their performance in tight spaces. 
For arbitrary shapes,~\cite{anyshape2025iros} achieves real-time planning using swept volume signed distance fields, though safety is enforced through penalty costs rather than hard constraints to limit additional complexity. Polytope-to-polytope obstacle avoidance CBF formulations have been proposed to address formal guarantees.~\cite{dcbf2022} uses a duality-based approach for an MPC-CBF framework, and~\cite{Chen2025cdcMinkowski} proposes a CBF constraint using the exact signed distance via Minkowski operations. These methods provide hard safety guarantees for non-circular robots, but they formulate avoidance constraints for each obstacle pair, scaling in complexity with the number of obstacles \cite{Li2024}. Moreover, they require obstacles to be represented as convex polytopes, whereas real environments are typically non-convex and described through occupancy grids or raw sensor data. Extracting a consistent set of polytopic obstacles is not always straightforward or  can become computationally heavy for large environments.\newline

\begin{figure}[tb]
    \centering
    \includegraphics[width=0.95\linewidth]{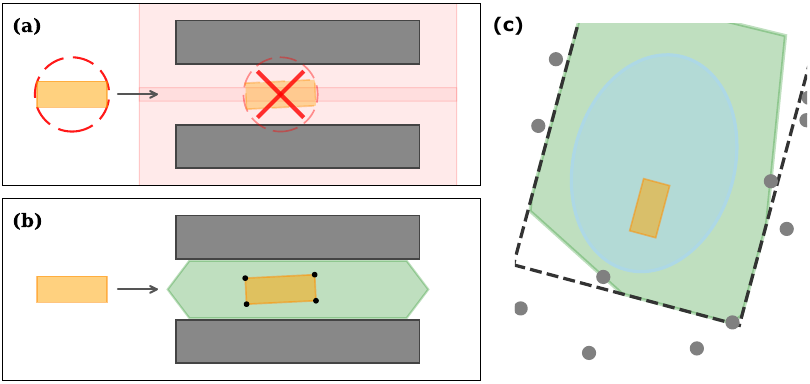}
    \caption{Circular footprint approximations do not permit to traverse through narrow spaces (a) where the robot should be able to pass based on its actual footprint (b).  Construction of the free-space polytope based on a heading-aligned bounding box (c).}
    \label{fig:conservatism}
\end{figure}

An alternative to obstacle-based formulations is to decompose the free space into convex regions. Methods such as IRIS~\cite{deits2015computing} and RILS~\cite{Liu2017PlanningEnvironments} compute convex polytopes directly from occupancy grids or point clouds that are guaranteed to be obstacle-free, without requiring individual obstacle identification and consideration. Recent advances have significantly improved the speed of these decompositions~\cite{werner-RSS-25}, while approaches such as FIRI~\cite{firi2025} additionally allow incorporating the robot's footprint as a seed, ensuring that the resulting region fully contains the robot geometry. These convex representations are well suited for optimization, as each region translates into a compact set of linear constraints. Trajectories through sequences of convex regions have been generated using optimization-based methods~\cite{deits2015icra,Marcucci2024,Marcucci-RSS-25, Scheffe2023} or joint optimization of trajectory and convex cover~\cite{wural2025}. Recently,~\cite{corridor2025cdc} proposed a CBF formulation to keep a robot within a free-space corridor, though the robot is assumed circular. As shown in Fig.~\ref{fig:conservatism}, this over-approximation for polytopic robots does not permit navigation through narrow spaces which are accessible considering the robot's actual geometry. In~\cite{Li2024}, containment of robots with general geometry inside convex free regions is addressed through polynomial positivity certificates and semidefinite programming, though the formulation optimizes a single-step control command rather than a receding-horizon trajectory. \newline

This work proposes a local motion planning and control method that guarantees to keep a polytopic robot inside convex free-space polytopes. It exploits free-space generation through FIRI, which produces convex polytopes guaranteed to contain the robot footprint.
A set of discrete CBF constraints inside an MPC problem guarantees the containment of the robot's polytopic footprint within the local free-space polytope over an horizon. Because safety is encoded against the free-space boundary rather than individual obstacles, the constraint complexity scales with the number of polytope hyperplanes rather than with the number of obstacles in the environment. The formulation is decoupled from the global reference generation and can be paired with any higher-level planner that provides a path or trajectory to follow. The main contributions are:
\begin{itemize}
\item A polytope-in-polytope CBF (PiP-CBF) formulation for MPC, guaranteeing that a convex polytopic footprint remains inside a local free-space region, 
\item A comparative analysis against a polytopic obstacle-based CBF method, demonstrating favorable scaling of the MPC problem size with increasing environmental complexity.
\item Experimental validation in a vessel simulation environment and a real-world autonomous mobile robot, constructing free-space regions from either prior occupancy grids or online LiDAR data, implicitly considering changes in the environment.
\end{itemize}

\section{Preliminaries}
\label{sec:preliminaries}

This section addresses the problem of motion planning with a polytopic robot inside an unstructured environment using a polytopic free-space decomposition. In addition, this section introduces required preliminaries on discrete-time CBFs for MPC.

\subsection{Problem Formulation}

Consider a robot operating in a planar environment $\mathcal{W} \subseteq \mathbb{R}^2$. The region occupied by obstacles is denoted $\mathcal{O} \subset \mathcal{W}$, and the free space is $\mathcal{F} = \mathcal{W} \setminus \mathcal{O}$. The environment is assumed to be represented by an occupancy grid or a set of points from a range sensor. No assumptions are made on any explicit obstacle decomposition. The robot's footprint is modeled as a convex polytope $\mathcal{P}_r$ with $n_r$ vertices, whose pose depends on the robot's state. A motion is considered safe if the footprint remains entirely within the free space, i.e., $\mathcal{P}_r(\boldsymbol{x}) \subseteq \mathcal{F}$, at all times. Since $\mathcal{F}$ is generally non-convex, enforcing this containment often renders the MPC hard and costly to solve. Instead, this work constructs a local convex inner approximation $\mathcal{P}_f \subseteq \mathcal{F}$ and reformulates the safety condition as $\mathcal{P}_r(\boldsymbol{x}) \subseteq \mathcal{P}_f$, which can be encoded into the MPC as a set of linear inequality constraints. The robot is assumed to receive approximate global information from a higher-level planner, and the objective is to follow this guidance while satisfying the safety condition.

\subsection{Discrete-Time Control Barrier Functions}

Consider a discrete-time system described by states $\boldsymbol{x} \in \mathcal{X} \subset \mathbb{R}^{n_x}$ and controls $\boldsymbol{u} \in \mathcal{U} \subset \mathbb{R}^{n_u}$, governed by
\begin{equation}
    \boldsymbol{x}_{k+1} = \boldsymbol{f}_d(\boldsymbol{x}_k, \boldsymbol{u}_k),
    \label{eq:dynamics}
\end{equation}
where $\boldsymbol{f}_d$ is locally Lipschitz. Let $h : \mathcal{X} \to \mathbb{R}$ be a continuous function, and define the associated safe set as
\begin{equation}
    \mathcal{C} = \{ \boldsymbol{x} \in \mathcal{X} \;|\; h(\boldsymbol{x}) \geq 0 \}.
    \label{eq:safe_set}
\end{equation}

\begin{definition}[Discrete-Time Control Barrier Function~\cite{dcbf2021}] \label{def:dcbf}
The function $h$ is a discrete-time control barrier function (DCBF) for system~\eqref{eq:dynamics} with respect to the set $\mathcal{C}$ if for all $\boldsymbol{x} \in \mathcal{C}$, there exists $\boldsymbol{u} \in \mathcal{U}$ such that
\begin{equation}
    h(\boldsymbol{f}_d(\boldsymbol{x}, \boldsymbol{u})) \geq (1 - \gamma)\, h(\boldsymbol{x}) \; , \quad 0 < \gamma \leq 1.
    \label{eq:dcbf}
\end{equation}
\end{definition}

The parameter $\gamma$ controls how aggressively the system may approach the boundary of $\mathcal{C}$: smaller values impose a slower decay of $h$, keeping the state further from the boundary.

\begin{proposition}[\cite{Agrawal2017}]
\label{prop:invariance}
If $h$ is a DCBF for system~\eqref{eq:dynamics}, $h(\boldsymbol{x}_0) \geq 0$, and a control input satisfying~\eqref{eq:dcbf} is applied at every time step, then $\boldsymbol{x}_k \in \mathcal{C}$ for all $k \geq 0$.
\end{proposition}

\subsection{Model Predictive Control with DCBF Constraints}

The DCBF condition~\eqref{eq:dcbf} can be enforced over a finite horizon within a MPC formulation~\cite{dcbf2021}. Over each control cycle, the following OCP is solved:
\begin{subequations}
\label{eq:mpc_dcbf}
\begin{align}
    \min_{\boldsymbol{x}_{k}, \boldsymbol{u}_{k}} \quad & \sum_{k=0}^{N-1} \ell(\boldsymbol{x}_{k}, \boldsymbol{u}_{k}) + \ell_{N}(\boldsymbol{x}_{N}) \label{eq:cost_function} \\
    \text{s.t.} \quad & \boldsymbol{x}_{0} = \hat{\boldsymbol{x}}, \label{eq:initial_condition} \\
    & \boldsymbol{x}_{k+1} = \boldsymbol{f}_d(\boldsymbol{x}_{k}, \boldsymbol{u}_{k}), \label{eq:dynamics_ocp} \\
    & \boldsymbol{x}_{k} \in \mathcal{X}, \quad \boldsymbol{u}_{k} \in \mathcal{U}, \label{eq:state_input_sets} \\
    & h(\boldsymbol{f}_d(\boldsymbol{x}_{l}, \boldsymbol{u}_{l})) \geq (1 - \gamma)\, h(\boldsymbol{x}_{l}), \label{eq:cbf_constraint} \\
    & l = 0, \ldots, N_{\text{cbf}} - 1, \nonumber \\
    & k = 0, \ldots, N - 1. \nonumber
\end{align}
\end{subequations}
where $\ell$ and $\ell_N$ are the stage and terminal costs, $\hat{\boldsymbol{x}}$ is the current state estimate, and $N_{\text{cbf}}$ is the horizon over which the safety constraint is enforced. Setting $1 < N_{\text{cbf}} < N$ leaves the remaining horizon stages free to progress  towards the goal while not impacting safety on the control sequence $\boldsymbol{u}_l$. The first element of the optimal control sequence $\boldsymbol{u}_0$ is applied, and the problem is re-solved at the next time step.


\section{Methodology}
\label{sec:methodology}
This section first describes the proposed framework for motion planning and control. Second, it details the proposed discrete-time CBF formulation to guarantee robot containment inside a free-space polytope. Last, it covers an inflation method to compute such free-space polytopes, including proposed heuristics to bias the inflation.

\subsection{Overview}

The proposed approach formulates safety as containment of the robot's polytopic footprint inside a convex free-space region, rather than as avoidance of individual obstacles. The environment is represented as a set of points, either extracted from an occupancy grid or obtained directly from a range sensor. From these points, a local convex polytope $\mathcal{P}_f \subseteq \mathcal{F}$ is constructed around the robot, and the MPC enforces a DCBF constraint that keeps the robot polytope inside this polytope throughout the safety horizon. A higher-level planner provides an approximate reference for the MPC to track. It is important to note that the containment formulation is a local safety layer and the MPC does not address global navigation. Without external global information, the robot may remain safely inside the current polytope indefinitely. The choice of higher-level planner is independent of the proposed method.

\subsection{Polytope-in-Polytope Control Barrier Function}
Instead of representing the environment as a set of obstacles, this paper proposes to model the local free-space as a $d$-dimensional convex polytope $\mathcal{P}_f$. A polytope $\mathcal{P}$ is fully described by a set of $n_{h}$ hyperplanes such that any point $\boldsymbol{p} \in \mathbb{R}^d$ inside the polytope satisfies
\begin{equation}
    \boldsymbol{W}\boldsymbol{p} + \boldsymbol{b} \geq 0,
    \label{eq:polytope}
\end{equation}
with $\boldsymbol{W} \in \mathbb{R}^{n_{h} \times d}$ and $\boldsymbol{b} \in \mathbb{R}^{n_{h}}$ being the normals and offsets of the hyperplanes, assumed to point inwards. Consider the robot modeled as a convex polytope $\mathcal{P}_{r}$ with $n_{r}$ vertices denoted by $\boldsymbol{v}^r_{i}$. It is fully contained inside $\mathcal{P}_f$ if and only if:
\begin{equation}
    \boldsymbol{W}^f\boldsymbol{v}^r_{i} + \boldsymbol{b}^f \geq 0 \quad \forall \,i \in [1, n_{r}].
\end{equation}

Let $\boldsymbol{w}^f_j \in \mathbb{R}^d$ be the $j$-th row of $\boldsymbol{W}^f$, representing the normal of the $j$-th hyperplane. For a robot at state $\boldsymbol{x}_{l}$, one can define the distance functions:
\begin{equation}
    h_{ij}(\boldsymbol{x}_{l}) = \boldsymbol{w}^f_j \cdot \boldsymbol{v}^r_i(\boldsymbol{x}_{l}) + b^f_{j} \label{eq:pip_cbf}
\end{equation}
for all $j \in [1, n_{h, f}]$ and $i \in [1, n_{r}]$. Enforcing $h_{ij}(\boldsymbol{x}_{l}) \geq 0$ on all vertices and hyperplanes guarantees containment of the robot inside $\mathcal{P}_f$. Subsequently, the discrete-time CBF condition~\eqref{eq:cbf_constraint} is formulated using all $n_r \times n_{h,f}$ functions $h_{ij}$:
\begin{equation}
    h_{ij}(\boldsymbol{f}_d(\boldsymbol{x}_{l}, \boldsymbol{u}_{l})) \geq (1 - \gamma)\, h_{ij}(\boldsymbol{x}_{l})
    \label{eq:pip_cbf_discrete}
\end{equation}
This ensures the robot's entire footprint remains safely contained within the free-space boundaries, strictly bounding the rate at which it may approach them.

\begin{remark}
For systems where the relative degree of the containment constraint with respect to the control input exceeds one, the PiP-CBF can be formulated as a discrete-time high-order CBF~\cite{DHOCBF}.
\end{remark}

\subsection{Free-Space Polytope Generation}
\label{sec:firi}

To construct the free-space polytope $\mathcal{P}_f$, the convex region inflation method FIRI~\cite{firi2025} is employed as visualized in Fig.~\ref{fig:conservatism}c. It produces a polytope, described by a set of hyperplanes, from a set of occupied points or polytopic obstacle descriptions, and a convex polygon as seed. The algorithm iteratively selects hyperplanes that separate the seed from nearby occupied regions while maximizing the volume of an inscribed ellipsoid. The resulting convex polytope is guaranteed to contain the seed without overlapping with occupied space. FIRI maximizes the inscribed ellipsoid volume, which produces the largest possible convex region around the seed.

In the proposed framework, the robot's rectangular footprint, slightly inflated by a safety margin, serves as the seed polygon. Occupied points are obtained either from occupancy grid cell corners or directly from a LiDAR point cloud. The polytope is regenerated at every control cycle to guarantee containment of the robot, and its hyperplanes $(\boldsymbol{W}^f, \boldsymbol{b}^f)$ enter the MPC formulation~\eqref{eq:mpc_dcbf} as constraint parameters through~\eqref{eq:pip_cbf}. Because the CBF constraint is only dependent on the free-space boundary, the framework does not require detecting, segmenting, or tracking individual obstacles. This improves uptake in real-world environments where obstacle decomposition is not trivial. Any environment representation that provides a set of occupied points is a valid input to the polytope generation.

\subsubsection*{Heuristic for Task-Informed Polytope Inflation}
To shape the polytope towards a region of interest, a bounding box is constructed around the robot, and the box's edges form the initial set of hyperplanes prior to ellipsoid inflation. This confines the inflation in large maps and biases the resulting polytope. The bounding box dimensions are a design choice that can reflect a preferred direction of motion. Additionally, 
the seed can incorporate both the current footprint and future footprints or waypoints, on the condition that the resulting seed remains convex and is trivially verifiable to be collision-free. In this work, a longer extent of the bounding box ahead of the robot and a shorter one behind is used, favoring forward navigation. 
The design of the bounding box or the choice of seed are heuristics. If the local geometry does not admit a polytope extending in a useful direction, the robot remains safe inside the current region but cannot progress toward the goal.

\section{Scalability Analysis}
\label{sec:scalability}
MPC on a real-world setup must adhere to strict computational time limits to match the required control frequency of the system. Hence, tractability of the MPC is of the utmost importance. State-of-the-art interior-point NLP solvers such as \texttt{Ipopt}~\cite{ipopt} and \texttt{Fatrop}~\cite{vanroye2023fatrop} solve a symmetric, indefinite perturbed Karush-Kuhn-Tucker (KKT) system at every iteration. The worst-case computational complexity to solve this system scales cubically with the number of variables and constraints per control interval~\cite{nocedal}. As such, keeping the problem size small and as convex as possible are two essential characteristics to keep this complexity low and adhere to limited computational resources. This section first discusses the theoretical scalability of the MPC dimensions for the PiP-CBF formulation compared to a benchmark method, with both methods assuming obstacles given as polytopes. Afterwards it provides quantitative validation of the practical implications of this scalability on extensive benchmark experiments.

\subsection{Theoretical Scalability}
The duality-based CBF (D-CBF) approach from~\cite{dcbf2022} is taken as the benchmark method. Table~\ref{tab:constraints_variables} summarizes the additional constraints and variables that both methods add to~\eqref{eq:mpc_dcbf}. To enforce safety,~\cite{dcbf2022} adds both variables and constraints to the MPC problem. On the other hand, PiP-CBF only adds the constraints through~\eqref{eq:pip_cbf}. In terms of their environmental complexity (individual obstacles vs free-space region), the number of variables ($N_x$) and constraints ($N_g$) added per control interval $[t_k, t_{k+1}]$ are:
\begin{align}
    \text{D-CBF}: \; &N_{x} + N_{g} = N_{o}(2n_{h, o} + 2n_r + 6)\label{eq:scalability_theory_dcbf}\\
    \text{PiP-CBF}: \; &N_{x} + N_{g} = n_{h, f}n_{r},
    \label{eq:scalability_theory_pip}
\end{align}
where $n_{h, o}$ and $n_{h, f}$ are the number of hyperplanes used to respectively represent an obstacle or the free-space and $N_{o}$ is the number of obstacles in the environment. As demonstrated in~\eqref{eq:scalability_theory_dcbf} and ~\eqref{eq:scalability_theory_pip}, PiP-CBF is independent of the number of obstacles and only depends on the geometric complexity of the free-space polytope. In contrast, D-CBF is dependent on both the environment and the obstacles' geometric complexity. Note that in cases with linear system dynamics, PiP-CBF requires only linear constraints, easing its adoption for (S)QP solvers.

\begin{figure}
    \centering
    \includegraphics[width=1.0\linewidth]{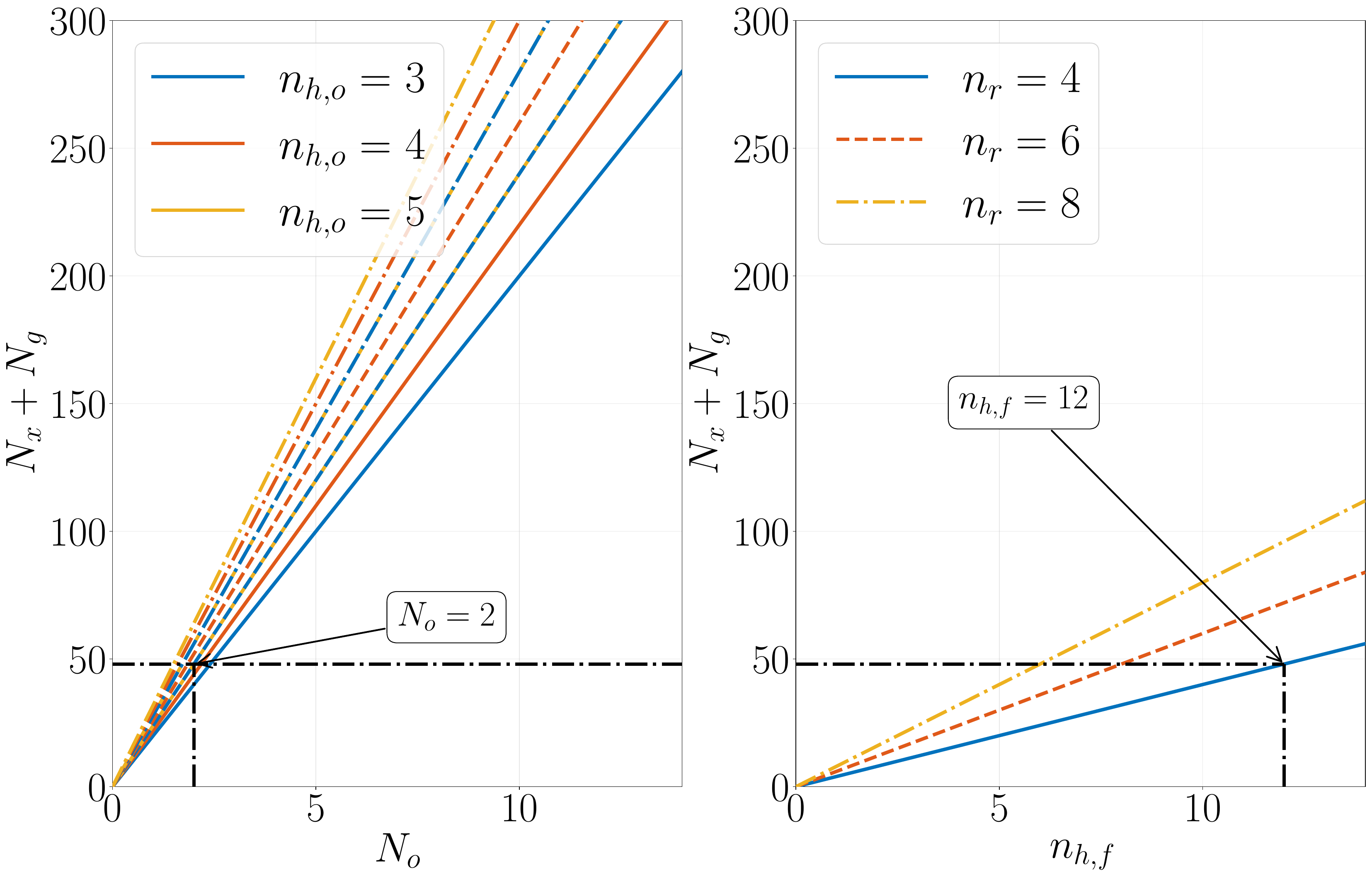}
    \caption{Theoretical comparison of the inflation of the MPC problem size for D-CBF (left) and PiP-CBF (right) in terms of their environmental complexity (number of obstacles vs. number of free-space hyperplanes). All full, dashed and dash-dotted lines correspond to the results for $n_{r}$ being four, six or eight, respectively.}
    \label{fig:scalability}
\end{figure}

Figure~\ref{fig:scalability} visualizes the number of extra constraints and variables per control interval for each method against its own measure of environmental complexity: the number of obstacles for D-CBF and the number of free-space hyperplanes for PiP-CBF. These relations clearly show that PiP-CBF scales favorably compared to D-CBF in terms of environmental complexity. D-CBF adds a significant amount of variables and constraints per control interval while PiP-CBF only adds a small amount of constraints, without a direct dependency on the number of obstacles. For a comparable problem size, PiP-CBF accommodates a free-space polytope of twelve hyperplanes, whereas D-CBF supports only two obstacles, illustrating how much more efficiently PiP-CBF captures environmental complexity. Applying PiP-CBF enables us to safely navigate through complex environments while keeping the MPC problem tractable.

\begin{table}[t]
\centering
\caption{D-CBF and PiP-CBF formulation per interval $[t_k, t_{k+1}]$.}
\label{tab:constraints_variables}
\setlength{\tabcolsep}{3pt}
\scriptsize
\begin{tabular}{c c c}
\toprule
  & Variables & Constraints\\
\midrule
\makecell[c]{D-CBF \\
$i \in [1, N_{o}]$} & $\begin{aligned}
      \lambda_{k}^{\mathcal{O}_{i}} &\in \mathbb{R}^{n_{h,o}} \nonumber \\ 
      \lambda_{k, i}^{\mathcal{R}} &\in \mathbb{R}^{n_{r}} \nonumber \\[-0.5ex]
      \omega_{k, i} &\in \mathbb{R} \nonumber
  \end{aligned}$ & $\begin{aligned}
    -\lambda_{k}^{\mathcal{O}_{i}}\boldsymbol{b}^{\mathcal{O}_{i}} - \lambda_{k, i}^{\mathcal{R}_{i}}\boldsymbol{b}^{\mathcal{R}} &\geq \omega_{k, i}(\Pi_{j=0}^k\gamma_{j})h_{i}(\boldsymbol{x}_{k})\\[-0.5ex]
    \|\lambda_{k}^{\mathcal{O}_{i}}\boldsymbol{W}^{\mathcal{O}_{i}}\|_{2}^{2} &\leq 1\\[-0.5ex]
    \lambda_{k}^{\mathcal{O}_{i}}\boldsymbol{W}^{\mathcal{O}_{i}} + \lambda_{k, i}^{\mathcal{R}}\boldsymbol{W}^{\mathcal{R}} &= 0 \\[-0.5ex]
    \lambda_{k}^{\mathcal{O}_{i}} \; ; \; \lambda_{k, i}^{\mathcal{R}} \; ; \; \omega_{k, i} &\geq 0
  \end{aligned}$\\
\midrule
\makecell[c]{PiP-CBF \\
$l \in [1, n_{r}]$} & / &
    \multicolumn{1}{l}{\hspace{5ex}$\boldsymbol{W}^{f}\boldsymbol{p}_l(\boldsymbol{x}_{k}) + \boldsymbol{b}^{f} \leq 0$} \\
\bottomrule
\end{tabular}
\end{table}


\begin{table*}[t]
\centering
\caption{Comparison D-CBF and PiP-CBF for a varying $N_{O}$.}
\label{tab:scalability_results}
\setlength{\tabcolsep}{9.5pt}
\scriptsize
\begin{tabular}{r c c c c c c c c c c c}
\toprule
Metric & Method & 1 & 2 & 3 & 4 & 5 & 6 & 7 & 8 & 9 & 10\\
\midrule
\multirow{2}{*}{Iterations [-]} & D-CBF & 7 & 9 & 9 & 10 & 13 & 14 & 14 & 15 & 16 & 16\\
 & PiP-CBF & \textbf{5} & \textbf{5} & \textbf{5} & \textbf{6} & \textbf{8} & \textbf{10} & \textbf{10} & \textbf{11} & \textbf{9} & \textbf{10}\\
\midrule
\multirow{2}{*}{$N_x$ [-]} & D-CBF
 & 393 & 602 & 811 & 1020 & 1129 & 1438 & 1647 & 1856 & 2065 & 2274 \\
 & PiP-CBF
 & \textbf{184} & \textbf{184} & \textbf{184} & \textbf{184} & \textbf{184} & \textbf{184} & \textbf{184} & \textbf{184} & \textbf{184} & \textbf{184} \\
 \midrule
\multirow{2}{*}{$N_g$ [-]} & D-CBF
 & \textbf{525} & 810 & 1095 & 1380 & 1665 & 1950 & 2235 & 2520 & 2805 & 3090 \\
 & PiP-CBF
 & 620 & \textbf{696} & \textbf{772} & \textbf{772} & \textbf{848} & \textbf{848} & \textbf{1000} & \textbf{1000} & \textbf{848} & \textbf{848} \\
 \midrule
 Max.~$n_{h, f}$ [-] & PiP-CBF
 & 5 & 6 & 7 & 7 & 8 & 8 & 10 & 10 & 8 & 8 \\
 \midrule
Reduction $t_{\text{wall}}$ [-] &  & 1.41$\times$ & 3.54$\times$ & 6.33$\times$ & 10.67$\times$ & 17.47$\times$ & 23.31$\times$ & 31.36$\times$ & 42.90$\times$ & 75.39$\times$ & 91.65$\times$\\
\bottomrule
\end{tabular}
\end{table*}

\subsection{Quantitative Comparison}\label{sec:quant}
\begin{figure}
    \centering
    \includegraphics[width=1.0\linewidth]{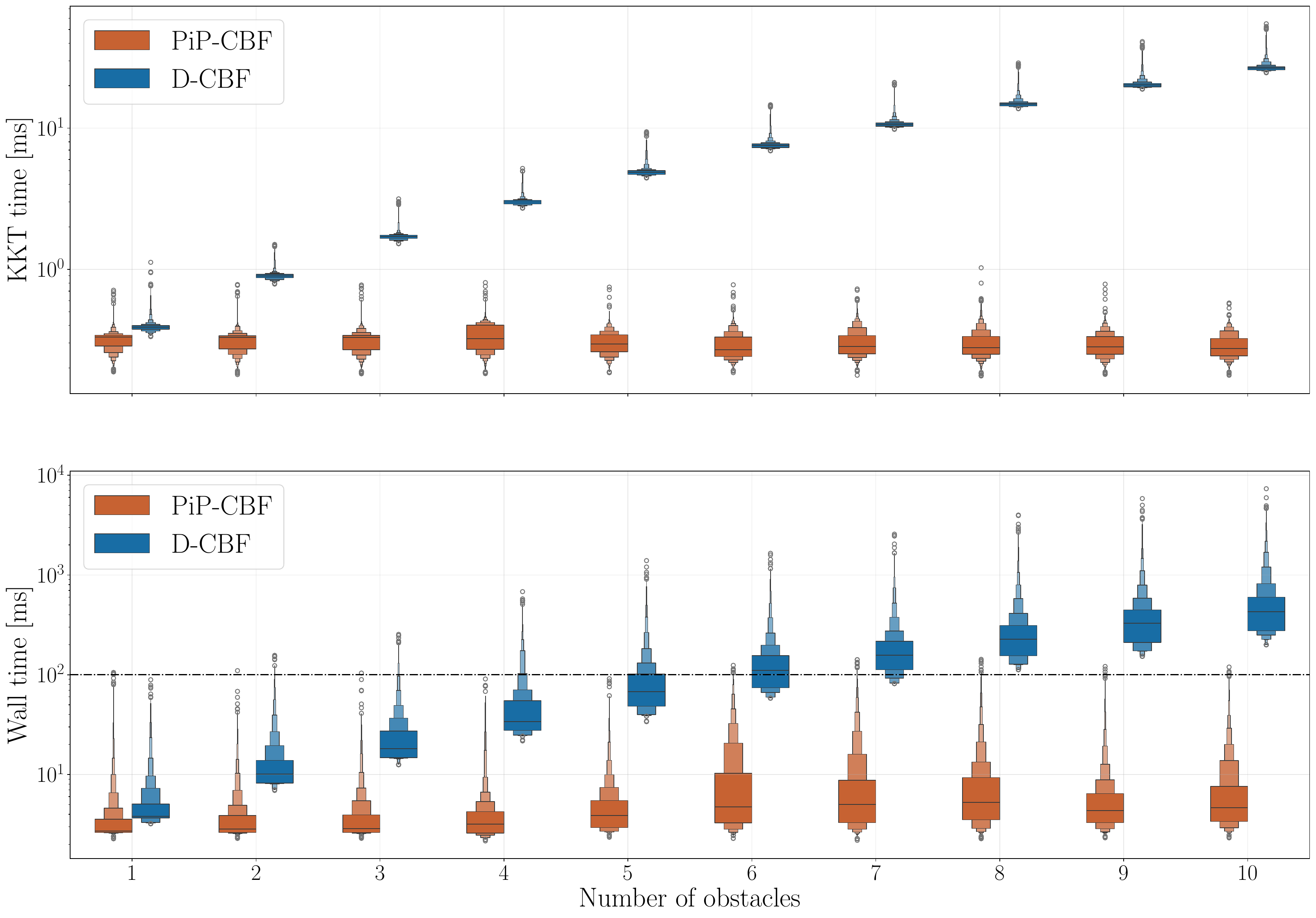}
    \caption{Quantitative validation for an increasing amount of obstacles of the scalability of D-CBF and PiP-CBF in terms of the wall time per iteration to solve the KKT system (top) and the total wall time to solve the MPC problem (bottom).}
    \label{fig:scalability_kkt_total}
\end{figure}

To quantify the practical implications of the increased problem dimensions, a scalability benchmark is conducted for both methods. Experiments are performed on a Nvidia Jetson Orin AGX using \texttt{Fatrop}~\cite{vanroye2023fatrop} through \texttt{CasADi}~\cite{Andersson2019} to solve the MPC problem. 100 environments, containing up to ten obstacles, are pseudo-randomly generated. Both methods have to navigate a rectangular robot through each environment between four distinct start and goal pose combinations, using \textit{A*}~\cite{astar} as global reference path. This results in more than 130,000 MPC solutions composing our scalability data. FIRI~\cite{firi2025} computes the free-space polytope for PiP-CBF using the robot footprint plus two points of the global path in front of the robot as seed. FIRI directly uses the polytopic obstacles as input. The horizons $N$ and  $N_{\text{cbf}}$ are chosen as 30 and 20, with $\gamma= 0.2$. Fig.~\ref{fig:scalability_kkt_total} reports the wall time per iteration spent in \texttt{Fatrop} to compute a new search direction (i.e., solve the KKT system, determine step size, etc.) and the total wall time to solve the MPC problem. Table~\ref{tab:scalability_results} reports the number of constraints and variables added to the MPC problem, the median number of iterations, the maximum number of free-space hyperplanes $n_{h, f}$ per number of obstacles, and the reduction of the median wall time required to solve the MPC problem. Across all generated environments, both methods achieve a 100\% success rate in reaching the goal while maintaining strict collision avoidance throughout the trajectory.

First, Table~\ref{tab:scalability_results} shows that PiP-CBF adds up to 12$\times$ fewer variables and 3.6$\times$ fewer constraints due to its independence of the number of obstacles. This effect is reflected in the time spent to compute a new search direction, shown at the top of Fig.~\ref{fig:scalability_kkt_total}. D-CBF's wall time increases cubically with the number of obstacles while PiP-CBF's wall time stays almost constant. The bottom figure shows a similar trend for the total computation time. The median wall time is reduced between $1.4\times$ and $91\times$ for environments up to ten obstacles. Both reductions are attributed to the limited increase in the complexity of the free-space polytope and thus the MPC problem dimensions in~\eqref{eq:scalability_theory_pip} for an increasingly cluttered environment. At a required control rate of 10~Hz, D-CBF remains tractable only up to four obstacles, beyond which its median wall time exceeds the 100~ms budget. Restricting D-CBF to the nearest obstacles could circumvent this limitation. However, the bottom row of Table~\ref{tab:scalability_results} shows that even with this restriction, PiP-CBF reduces computation time by $1.4\times$ to $10.67\times$. This improvement allows to: i) increase the control frequency or horizon, ii) control more complex systems or iii) perform other necessary computations. Second, PiP-CBF requires between 7\% and 44\% fewer iterations at the median to solve the MPC problem. This is due to PiP-CBF's convex structure, compared to D-CBF's non-convex formulation, in turn also contributing to lower total computation times.


\section{Experimental Validation}
\label{sec:experiments}
This section addresses the experimental validation of the proposed framework, including simulation experiments with an autonomous surface vessel, and real-world experiments with an autonomous mobile robot prototype. Furthermore, a discussion of the benefits and limitations is presented.

\begin{figure}[tb]
    \centering    \includegraphics[width=0.95\linewidth] {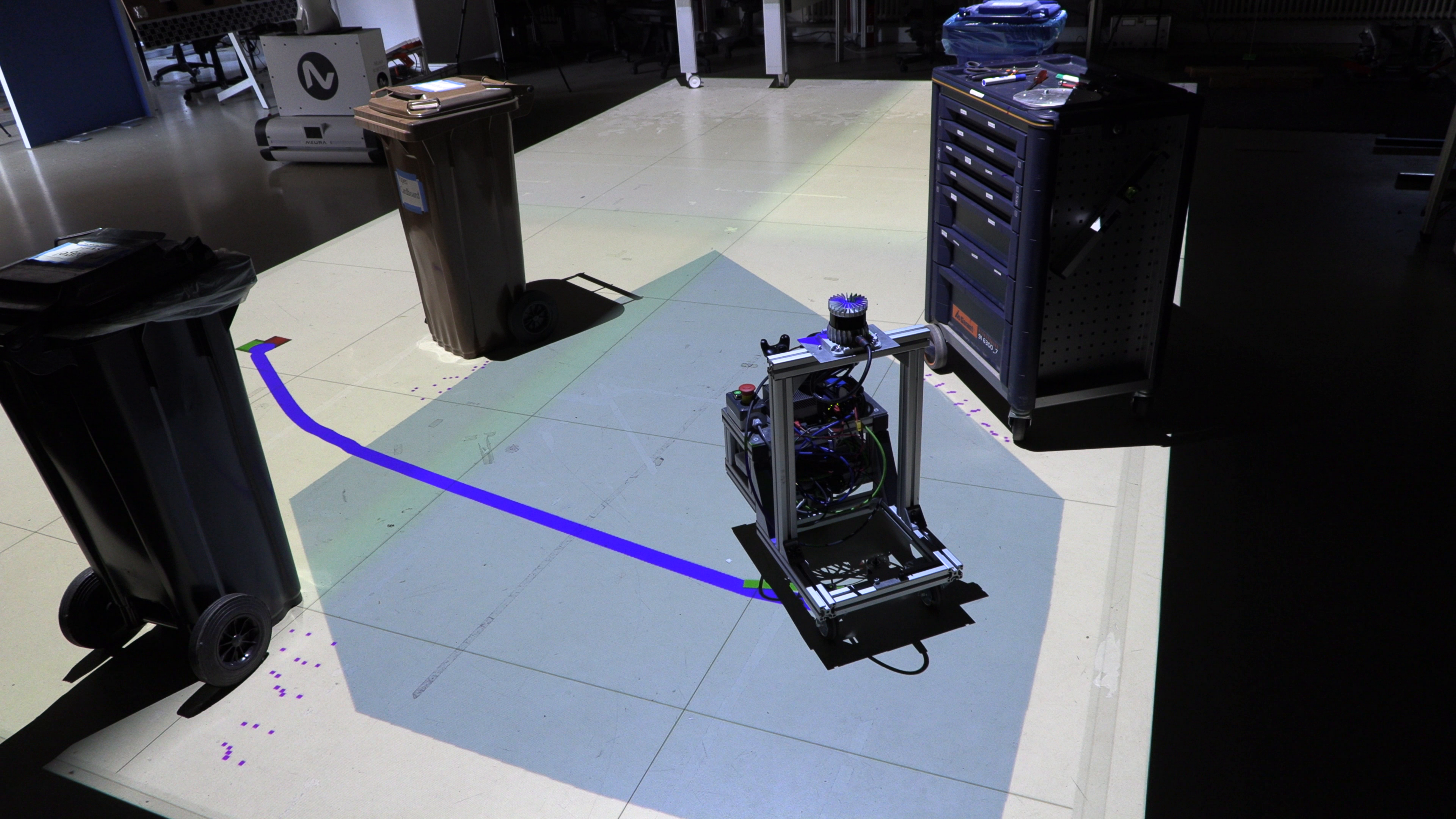}
    \caption{Visualization of the real-world lab setup including the AMR platform, the constructed free-space polytope based on the LiDAR's point cloud and the global A* reference.}
    \label{fig:amr_photo}
\end{figure}

\begin{figure}[tb]
    \centering
    \includegraphics[width=0.85\linewidth]{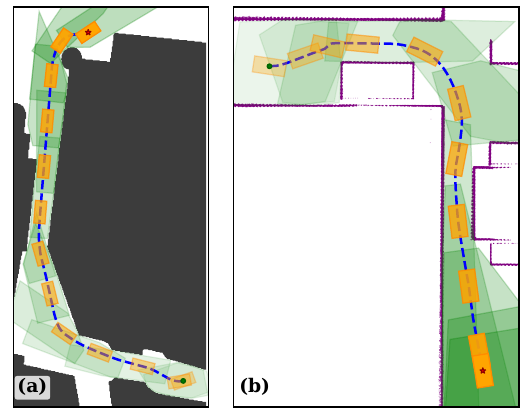}
    \vspace{1em} 
    \includegraphics[width=0.85\linewidth]{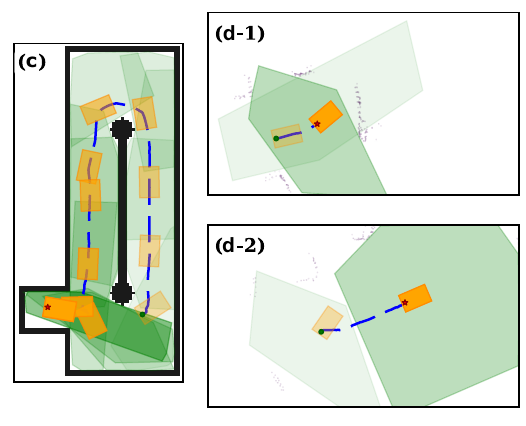}
    \caption{Illustration of runs for the ASV (a, b) and AMR (c, d) cases using occupancy grids (a, c) and LiDAR (b, d). The traveled path is visualized in blue, free-space polytopes at various time instances are indicated in green. (d-1) and (d-2) depict different time instances of a dynamic environment showing the AMR safely navigating around a dynamic obstacle.}
    \label{fig:sim_environments}
\end{figure}

\begin{figure} [t]
    \centering
    \includegraphics[width=1.0\linewidth]{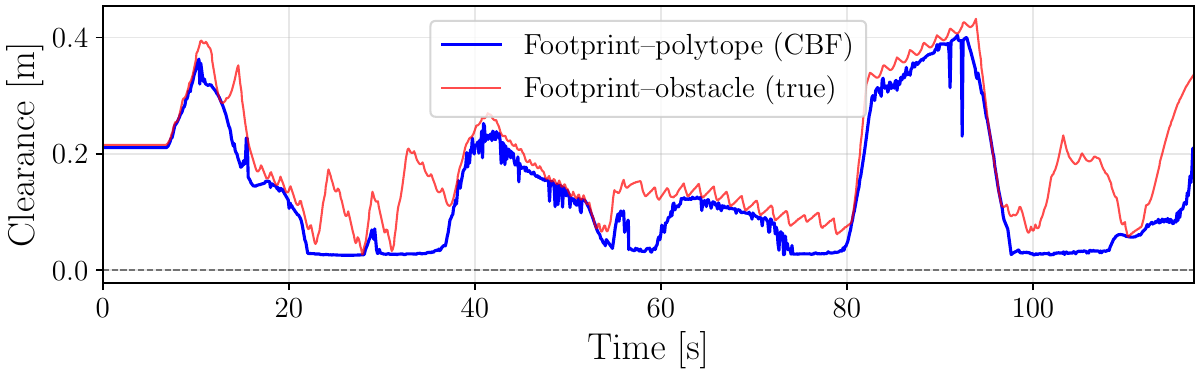}
    \caption{Visualization of PiP-CBF's conservatism for the environment in Fig.~\ref{fig:sim_environments}a in terms of the clearance to the edge of the free-space polytope (blue) and the actual obstacles (red).}
    \label{fig:conservatism_ts}
\end{figure}

\subsection{Setup}

The framework is validated in simulation with an Autonomous Surface Vehicle (ASV) and on hardware with an Autonomous Mobile Robot (AMR) with bicycle kinematics. For each case, two sets of experiments are conducted: i) using static occupancy grids, and ii) using LiDAR sensing. These configurations exercise distinct aspects of the framework. The ASV and AMR platforms test the approach under holonomic and non-holonomic dynamics respectively. The occupancy grid and scan configurations correspond to two standard operating regimes: a known environment, where a global path can be computed on a static map, and a reactive, unknown setting where the free space is constructed solely from online sensing. The LiDAR configuration additionally lets us evaluate performance in dynamic environments, with slowly moving obstacles that are not part of any prior map. All experiments run on a NVIDIA Jetson Orin AGX, with a desired update rate of 10~Hz. The MPC is solved using \texttt{Fatrop}~\cite{vanroye2023fatrop} through \texttt{CasADi}~\cite{Andersson2019}. The prediction horizon is $N=40$ for the ASV and $N=20$ for the AMR, with $N_{\text{cbf}}=10$ for ASV grid map runs, $N_{\text{cbf}}=20$ for ASV LiDAR runs, and $N_{\text{cbf}}=18$ for all AMR runs. The CBF decay rate is $\gamma = 0.2$, and the OCP accepts a maximum of 20 hyperplanes per polytope.

The ASV simulator follows the dynamics of~\cite{WeiICRA2018}, implemented in ROS~\cite{Quigley09}. The grid map configuration covers four segments based on Amsterdam canals, with \textit{A*} providing global guidance and polytopes are generated directly from the occupancy grid. The LiDAR configuration uses Gazebo~\cite{gazebo} to simulate sensor data in three cluttered canal intersections, computes an approximate plan via lexicographic search~\cite{Shan2020}, and builds a polytope using only current laser scan data.

The AMR prototype (Fig.~\ref{fig:amr_photo}) uses a KELO Drive 100 as a steerable front wheel, with a wheelbase of $L_0 = 0.42$~m and a $0.55 \times 0.35$~m rectangular footprint. Localization is provided by an HTC VIVE Tracker 3.0 with four base stations. An Ouster OS0 LiDAR is mounted on the robot and used for obstacle sensing in the LiDAR configuration. The full AMR software stack is built on ROS 2~\cite{macenski2022robot}. For each run, manual goal positions are issued, and \textit{A*} computes an approximate reference path in both configurations. For grid map runs, both the global path and polytopes are built online with the static maps across ten indoor scenarios. For LiDAR runs, the robot accumulates scans during navigation into an online map for \textit{A*} guidance while polytopes are generated from the point cloud data; these experiments include scenarios with slowly moving dynamic obstacles. Environmental features such as the occupancy grid, the point clouds and the free-space polytope are visualized as projections onto the laboratory floor.

\subsection{Results \& Discussion}

Table~\ref{tab:results} summarizes the MPC wall time and polytope statistics across all four experiment configurations. Representative trajectories are shown in Fig.~\ref{fig:sim_environments}, and the supplementary video shows runs from each configuration.

\begin{table}[tb]
\centering
\caption{MPC with PiP-CBF solve time and polytope
statistics.}
\label{tab:results}
\setlength{\tabcolsep}{3.25pt}
\small
\begin{tabular}{l c c c c c c c}
\toprule
& \multirow{2}{*}[-0.4ex]{Runs}
& \multicolumn{3}{c}{$t_\text{wall}$ [ms]}
& \multicolumn{3}{c}{$n_{h,f}$} \\
\cmidrule(lr){3-5} \cmidrule(lr){6-8}
& & median & p99 & max
& mean & range & ${>}15$ [\%] \\
\midrule
ASV (map)   & 4  & 5.5  & 10.6  & 100.9 & 9.9  & 5--20 & 3.2 \\
ASV (scan) & 3  & 11.1 & 38.6  & 100.8 & 9.9  & 7--16 & 0.2 \\
AMR (map)   & 10 & 34.2 & 79.8  & 91.5  & 10.5 & 7--17 & 0.5 \\
AMR (scan) & 15 & 46.4 & 89.4  & 124.3 & 11.3 & 7--16 & 0.1 \\
\bottomrule
\end{tabular}
\end{table}

In the ASV simulations, fewer than 0.04\% of computation times exceeded the desired 100~ms control period and no collisions were observed. The number of active hyperplanes averaged around 10 in both sensing modes. This confirms that, independent of the sensor input, the polytope complexity is governed by the local geometry.
Since the free-space polytope is a convex inner approximation of the true (generally non-convex) obstacle-free region, it necessarily discards some navigable space. Fig.~\ref{fig:conservatism_ts} illustrates this effect for the ASV occupancy grid run from Fig.~\ref{fig:sim_environments}a by showing the time evolution of the minimum distance from the footprint to the polytope boundary alongside the minimum distance from the footprint to the nearest occupied point. Near bends and intersections, where the free space is highly non-convex, the polytope boundary is closer to the footprint than the actual obstacles, reflecting the space lost by the convex approximation. In tight passages, where the obstacles are the binding constraint, both distances converge and the approximation is near-exact. In Fig.~\ref{fig:conservatism_ts}, the minimum footprint-to-polytope distance was 2.6~cm and the minimum footprint-to-obstacle distance 2.7~cm, both occurring at the tightest point of the trajectory.

On the AMR case, computation times were higher due to the combination of the complex kinematics with the steerable front wheel and the navigation through narrow spaces. The median and 99th percentile nevertheless remained below the desired 100~ms budget in both configurations, with fewer than 0.1\% of outliers above the threshold. No collisions were observed across any run, including the ten AMR LiDAR runs in which humans (pushing carts in some cases) crossed the workspace during navigation. Fig.~\ref{fig:sim_environments}c shows the AMR executing a reverse parking maneuver, where the footprint always remains inside the generated polytopes throughout the turn. Fig.~\ref{fig:sim_environments}d shows the polytope contracting as a dynamic obstacle enters the previously constructed polytope and re-expanding after it passes.  

Setting $N_{\text{cbf}} < N$ allows the prediction horizon to extend beyond the current free-space polytope. This guides the robot toward the goal and avoids explicit corridor sequencing. Our bounding box heuristic biases the convex coverage toward the robot's forward direction, and produced a polytope covering a feasible path at every replanning cycle across the 32 reported runs. The heuristic is a tuning choice rather than a guarantee of navigation completeness: when the feasible path requires motion outside the box, or when the reference itself is infeasible, the robot stops safely but cannot make progress. Generating convex free-space regions that satisfy downstream navigation requirements is not a trivial task~\cite{firi2025}, and coupling the construction of the bounding box to the reference path or to the robot's dynamics is a direction for future work.

Across all experiments, the free-space hyperplane count remained between 5 and 20 regardless of the environment or its representation, with fewer than 4\% exceeding 15 in any configuration. The framework operated without structural modification to the MPC or CBF formulation across holonomic and non-holonomic dynamics; grid map and LiDAR inputs; and simulated and physical environments. No collisions were observed in any of the 32 reported runs. Although dynamic obstacles are not explicitly considered in the formulation, recomputing the polytope at every time step allows the method to directly react to changes in the environment, including slowly moving obstacles or people.

\section{Conclusions \& Future Work}

This work presented a polytope-in-polytope MPC-CBF formulation that keeps a convex robot footprint inside a free-space polytope. Because safety is encoded against the free-space boundary rather than against individual obstacles, the constraint complexity scales with the number of free-space hyperplanes and robot vertices, independent of the number of obstacles in the environment. This also eliminates the need for obstacle segmentation or tracking in the perception pipeline: the controller operates directly on occupancy grids or point clouds, and is compatible with any reference source that provides a path or goal position. The approach is validated across holonomic and non-holonomic dynamics, grid map and LiDAR inputs, and simulated and physical environments. All computations are executed online on an onboard embedded platform. Future work includes coupling the polytope generation to the robot's dynamics or the planned trajectory, so that the free-space region accounts for future robot navigation instead of only the current status. This will improve safety and overall smoothness as consecutive polytopes account for the robot's behaviour over subsequent control inputs. The polytope construction will inherently follow the robot's intended motion direction, removing the need for a heuristic to progress towards the goal. Involving the (estimated) motion of dynamic obstacles into the construction of (a sequence) of free-space regions forms an interesting future research direction. Formal guarantees on corridor continuity over consequent MPC solutions and navigation completeness remain open challenges.




\bibliographystyle{IEEEtran}
\bibliography{References}

@ARTICLE{Marcucci2024,
  author={Marcucci, Tobia and Nobel, Parth and Tedrake, Russ and Boyd, Stephen},
  journal={IEEE Transactions on Robotics}, 
  title={{Fast Path Planning Through Large Collections of Safe Boxes}}, 
  year={2024},
  volume={40},
  number={},
  pages={3795-3811},
  doi={10.1109/TRO.2024.3434168}}

@INPROCEEDINGS{Fiorini1993,

  author={Fiorini, P. and Shiller, Z.},

  booktitle={[1993] Proceedings IEEE International Conference on Robotics and Automation}, 

  title={Motion planning in dynamic environments using the relative velocity paradigm}, 

  year={1993},

  volume={},

  number={},

  pages={560-565 vol.1},

  doi={10.1109/ROBOT.1993.292038}}

@INPROCEEDINGS{Marcucci-RSS-25, 
    AUTHOR    = {Marcucci, Tobia and Halm, Mathew and Yang, William and Lee, Dongchan and Marchese, Andrew D}, 
    TITLE     = {{A Biconvex Method for Minimum-Time Motion Planning Through Sequences of Convex Sets}}, 
    BOOKTITLE = {Proceedings of Robotics: Science and Systems}, 
    YEAR      = {2025}
}

@book{nocedal,
edition = {2nd ed. 2006.},
language = {eng},
isbn = {0-387-40065-6},
publisher = {Springer New York},
series = {Springer Series in Operations Research and Financial Engineering},
title = {Numerical Optimization },
year = {2006},
author = {Nocedal, Jorge and Wright, Stephen},
address = {New York, NY},
booktitle = {Numerical Optimization}
}

@InProceedings{Khatib1985,
author = {Khatib, O},
booktitle = {Proceedings - IEEE International Conference on Robotics and Automation},
copyright = {Copyright 2018 Elsevier B.V., All rights reserved.},
isbn = {0818606150},
issn = {1050-4729},
pages = {500-505},
title = {Real-time obstacle avoidance for manipulators and mobile robots},
volume = {2},
year = {1985},
}

@ARTICLE{astar,
  author={Hart, Peter E. and Nilsson, Nils J. and Raphael, Bertram},
  journal={IEEE Transactions on Systems Science and Cybernetics}, 
  title={{A Formal Basis for the Heuristic Determination of Minimum Cost Paths}}, 
  year={1968},
  volume={4},
  number={2},
  pages={100-107},
  doi={10.1109/TSSC.1968.300136}}

@article{dijkstra1959note,
  title={{A Note on Two Problems in Connexion with Graphs.}},
  author={Dijkstra, E. W.},
  journal={Numerische Mathematik},
  volume={1},
  pages={269--271},
  year={1959}
}

@ARTICLE{occupancygrid,
  author={Elfes, A.},
  journal={Computer}, 
  title={{Using occupancy grids for mobile robot perception and navigation}}, 
  year={1989},
  volume={22},
  number={6},
  pages={46-57},
  doi={10.1109/2.30720}}

@ARTICLE{wural2025,
  author={Wu, Yuwei and Spasojevic, Igor and Chaudhari, Pratik and Kumar, Vijay},
  journal={IEEE Robotics and Automation Letters}, 
  title={{Towards Optimizing a Convex Cover of Collision-Free Space for Trajectory Generation}}, 
  year={2025},
  volume={10},
  number={5},
  pages={4762-4769},
  doi={10.1109/LRA.2025.3553416}}

@article{Liu2017PlanningEnvironments,
    title = {{Planning dynamically feasible trajectories for quadrotors using safe flight corridors in 3-D complex environments}},
    year = {2017},
    journal = {IEEE Robotics and Automation Letters},
    author = {Liu, Sikang and Watterson, Michael and Mohta, Kartik and Sun, Ke and Bhattacharya, Subhrajit and Taylor, Camillo J. and Kumar, Vijay},
    number = {3},
    month = {7},
    pages = {1688--1695},
    volume = {2},
    publisher = {Institute of Electrical and Electronics Engineers Inc.},
    doi = {10.1109/LRA.2017.2663526},
    issn = {23773766}
}

@inproceedings{deits2015computing,
  title={{Computing large convex regions of obstacle-free space through semidefinite programming}},
  author={Deits, Robin and Tedrake, Russ},
  booktitle={Algorithmic Foundations of Robotics XI: Selected Contributions of the Eleventh International Workshop on the Algorithmic Foundations of Robotics},
  year={2015},
}

@INPROCEEDINGS{deits2015icra,
  author={Deits, Robin and Tedrake, Russ},
  booktitle={2015 IEEE International Conference on Robotics and Automation (ICRA)}, 
  title={{Efficient mixed-integer planning for UAVs in cluttered environments}}, 
  year={2015},
  volume={},
  number={},
  pages={42-49},
}

@ARTICLE{firi2025,
  author={Wang, Qianhao and Wang, Zhepei and Wang, Mingyang and Ji, Jialin and Han, Zhichao and Wu, Tianyue and Jin, Rui and Gao, Yuman and Xu, Chao and Gao, Fei},
  journal={IEEE Transactions on Robotics}, 
  title={{Fast Iterative Region Inflation for Computing Large 2-D/3-D Convex Regions of Obstacle-Free Space}}, 
  year={2025},
  volume={41},
  number={},
  pages={3223-3243},
  doi={10.1109/TRO.2025.3562482}}

@INPROCEEDINGS{werner-RSS-25, 
    author={Werner, Peter and Cheng, Richard and Stewart, Tom and Tedrake, Russ and Rus, Daniela}, 
    title={{Superfast configuration-space convex set computation on {GPU}s for online motion planning}}, 
    BOOKTITLE = {Proceedings of Robotics: Science and Systems}, 
    YEAR      = {2025}
}

@article{macenski2022robot,
  title={Robot operating system 2: Design, architecture, and uses in the wild},
  author={Macenski, Steven and Foote, Tully and Gerkey, Brian and Lalancette, Chris and Woodall, William},
  journal={Science robotics},
  volume={7},
  number={66},
  year={2022},
  publisher={American Association for the Advancement of Science}
}

@InProceedings{macenski2020marathon2,
author = {Macenski, Steven and Martin, Francisco and White, Ruffin and Ginés Clavero, Jonatan},
title = {{The Marathon 2: A Navigation System}},
booktitle = {2020 IEEE/RSJ International Conference on Intelligent Robots and Systems (IROS)},
year = {2020}
}

@Article{Andersson2019,
  author = {Joel A E Andersson and Joris Gillis and Greg Horn
            and James B Rawlings and Moritz Diehl},
  title = {{{CasADi} -- {A} software framework for nonlinear optimization
           and optimal control}},
  journal = {Mathematical Programming Computation},
  volume = {11},
  number = {1},
  pages = {1--36},
  year = {2019},
  publisher = {Springer},
  doi = {10.1007/s12532-018-0139-4}
}

@inproceedings{vanroye2023fatrop,
  title={Fatrop: A fast constrained optimal control problem solver for robot trajectory optimization and control},
  author={Vanroye, Lander and Sathya, Ajay and De Schutter, Joris and Decr{\'e}, Wilm},
  booktitle={2023 IEEE/RSJ International Conference on Intelligent Robots and Systems (IROS)},
  year={2023},
}

@INPROCEEDINGS{anyshape2025iros,
  author={Wang, Yijin and Zhang, Tingrui and Zhang, Mengke and Ji, Shuhang and Li, Xiaoying and Gao, Fei},
  booktitle={2025 IEEE/RSJ International Conference on Intelligent Robots and Systems (IROS)}, 
  title={Any-shape Real-time Replanning via Swept Volume SDF}, 
  year={2025},
  volume={},
  number={},
  pages={2068-2075},
  doi={10.1109/IROS60139.2025.11246875}}

@INPROCEEDINGS{dcbf2022,
  author={Thirugnanam, Akshay and Zeng, Jun and Sreenath, Koushil},
  booktitle={2022 International Conference on Robotics and Automation (ICRA)}, 
  title={Safety-Critical Control and Planning for Obstacle Avoidance between Polytopes with Control Barrier Functions}, 
  year={2022},
  volume={},
  number={},
  pages={286-292},
  doi={10.1109/ICRA46639.2022.9812334}}

@INPROCEEDINGS{liu2025acc,
  author={Liu, Shuo and Mao, Yihui and Belta, Calin A.},
  booktitle={2025 American Control Conference (ACC)}, 
  title={Safety-Critical Planning and Control for Dynamic Obstacle Avoidance Using Control Barrier Functions}, 
  year={2025},
  volume={},
  number={},
  pages={348-354},
  doi={10.23919/ACC63710.2025.11107805}}

@INPROCEEDINGS{dcbf2021,
  author={Zeng, Jun and Zhang, Bike and Sreenath, Koushil},
  booktitle={2021 American Control Conference (ACC)}, 
  title={Safety-Critical Model Predictive Control with Discrete-Time Control Barrier Function}, 
  year={2021},
  volume={},
  number={},
  pages={3882-3889},
  doi={10.23919/ACC50511.2021.9483029}}

@INPROCEEDINGS{corridor2025cdc,
  author={Mohammad, Nicholas and Bezzo, Nicola},
  booktitle={2025 IEEE 64th Conference on Decision and Control (CDC)}, 
  title={Corridor-based Adaptive Control Barrier \& Lyapunov Functions for Safe Mobile Robot Navigation}, 
  year={2025},
  volume={},
  number={},
  pages={3107-3114},
  doi={10.1109/CDC57313.2025.11312266}}

@article{xiao2021high,
  title={High-order control barrier functions},
  author={ W. {Xiao} and C. {Belta}},
  journal={IEEE Transactions on Automatic Control},
  volume={67},
  number={7},
  pages={3655--3662},
  year={2021},
  publisher={IEEE}
}

@inproceedings{
WeiICRA2018,
   Author = {Wang, Wei and Mateos, Luis and Park, Shinkyu and Leoni, Pietro and Gheneti, Banti and Duarte, Fabio and Ratti, Carlo and Rus, Daniela},
   Title = {Design, Modeling, and Nonlinear Model Predictive Tracking Control of a Novel Autonomous Surface Vehicle},
   booktitle = {Proc. 2018 IEEE Int. Conf. Robot. Autom},
    pages={6189-6196},
    Year = {2018}
   }

@INPROCEEDINGS{Shan2020,
  author={Shan, Tixiao and Wang, Wei and Englot, Brendan and Ratti, Carlo and Rus, Daniela},
  booktitle={2020 59th IEEE Conference on Decision and Control (CDC)}, 
  title={A Receding Horizon Multi-Objective Planner for Autonomous Surface Vehicles in Urban Waterways}, 
  year={2020},
  volume={},
  number={},
  pages={4085-4092},
  doi={10.1109/CDC42340.2020.9304298}}

@inproceedings{Quigley09,
author={Morgan Quigley and Brian Gerkey and Ken Conley and Josh Faust and
Tully Foote and Jeremy Leibs and Eric Berger and Rob Wheeler and Andrew Ng},
title={{ROS}: an open-source Robot Operating System},
booktitle={Proc. of the IEEE Intl. Conf. on Robotics and Automation (ICRA)
Workshop on Open Source Robotics},
year={2009},
}

@INPROCEEDINGS{Chen2025cdcMinkowski,
  author={Chen, Yi-Hsuan and Liu, Shuo and Xiao, Wei and Belta, Calin and Otte, Michael},
  booktitle={2025 IEEE 64th Conference on Decision and Control (CDC)}, 
  title={Control Barrier Functions via Minkowski Operations for Safe Navigation among Polytopic Sets}, 
  year={2025},
  volume={},
  number={},
  pages={4481-4488},
  doi={10.1109/CDC57313.2025.11312188}}

@INPROCEEDINGS{conecbf2024acc,
  author={Tayal, Manan and Singh, Rajpal and Keshavan, Jishnu and Kolathaya, Shishir},
  booktitle={2024 American Control Conference (ACC)}, 
  title={Control Barrier Functions in Dynamic UAVs for Kinematic Obstacle Avoidance: A Collision Cone Approach}, 
  year={2024},
  volume={},
  number={},
  pages={3722-3727},
  doi={10.23919/ACC60939.2024.10644548}}

@INPROCEEDINGS{10886807,
  author={Gao, Yunfan and Messerer, Florian and Duijkeren, Niels van and Houska, Boris and Diehl, Moritz},
  booktitle={2024 IEEE 63rd Conference on Decision and Control (CDC)}, 
  title={Real-Time-Feasible Collision-Free Motion Planning For Ellipsoidal Objects}, 
  year={2024},
  volume={},
  number={},
  pages={5108-5113},
  doi={10.1109/CDC56724.2024.10886807}}

@INPROCEEDINGS{panoc2018ecc,
  author={Sathya, Ajay and Sopasakis, Pantelis and Van Parys, Ruben and Themelis, Andreas and Pipeleers, Goele and Patrinos, Panagiotis},
  booktitle={2018 European Control Conference (ECC)}, 
  title={Embedded nonlinear model predictive control for obstacle avoidance using PANOC}, 
  year={2018},
  volume={},
  number={},
  pages={1523-1528},
  doi={10.23919/ECC.2018.8550253}}

@ARTICLE{508439,
  author={Kavraki, L.E. and Svestka, P. and Latombe, J.-C. and Overmars, M.H.},
  journal={IEEE Transactions on Robotics and Automation}, 
  title={{Probabilistic roadmaps for path planning in high-dimensional configuration spaces}}, 
  year={1996},
  volume={12},
  number={4},
  pages={566-580},
  doi={10.1109/70.508439}}

@ARTICLE{11282962,
  author={Mendes, João Félix and Basiri, Meysam and Ventura, Rodrigo},
  journal={IEEE Robotics and Automation Letters}, 
  title={Kinodynamic Trajectory Planning for Efficient UAV Exploration and Reconstruction of Unknown Environments}, 
  year={2026},
  volume={11},
  number={2},
  pages={1530-1537},
  doi={10.1109/LRA.2025.3641147}}

@article{MAYNE2000789,
title = {Constrained model predictive control: Stability and optimality},
journal = {Automatica},
volume = {36},
number = {6},
pages = {789-814},
year = {2000},
issn = {0005-1098},
doi = {https://doi.org/10.1016/S0005-1098(99)00214-9},
author = {D.Q. Mayne and J.B. Rawlings and C.V. Rao and P.O.M. Scokaert}
}

@ARTICLE{11419776,
  author={Shahna, Mehdi Heydari and Mustalahti, Pauli and Mattila, Jouni},
  journal={IEEE Robotics and Automation Letters}, 
  title={NMPC-Augmented Visual Navigation and Safe Learning Control for Large-Scale Mobile Robots}, 
  year={2026},
  volume={11},
  number={4},
  pages={5182-5189},
  doi={10.1109/LRA.2026.3669802}}

@ARTICLE{9385847,
  author={Brito, Bruno and Everett, Michael and How, Jonathan P. and Alonso-Mora, Javier},
  journal={IEEE Robotics and Automation Letters}, 
  title={Where to go Next: Learning a Subgoal Recommendation Policy for Navigation in Dynamic Environments}, 
  year={2021},
  volume={6},
  number={3},
  pages={4616-4623},
  doi={10.1109/LRA.2021.3068662}}

@inproceedings{ames2019cbftheory,
  title={Control barrier functions: Theory and applications},
  author={Ames, Aaron D and Coogan, Samuel and Egerstedt, Magnus and Notomista, Gennaro and Sreenath, Koushil and Tabuada, Paulo},
  booktitle={2019 18th European control conference (ECC)},
  pages={3420--3431},
  year={2019},
  organization={Ieee}
}

@INPROCEEDINGS{gazebo,
  author={Koenig, N. and Howard, A.},
  booktitle={2004 IEEE/RSJ International Conference on Intelligent Robots and Systems (IROS) (IEEE Cat. No.04CH37566)}, 
  title={Design and use paradigms for Gazebo, an open-source multi-robot simulator}, 
  year={2004},
  volume={3},
  number={},
  pages={2149-2154 vol.3},
  doi={10.1109/IROS.2004.1389727}}

@ARTICLE{Scheffe2023,
  author={Scheffe, Patrick and Henneken, Theodor Mario and Kloock, Maximilian and Alrifaee, Bassam},
  journal={IEEE Transactions on Intelligent Vehicles}, 
  title={Sequential Convex Programming Methods for Real-Time Optimal Trajectory Planning in Autonomous Vehicle Racing}, 
  year={2023},
  volume={8},
  number={1},
  pages={661-672},
  doi={10.1109/TIV.2022.3168130}}

@article{ipopt,
  title={On the implementation of an interior-point filter line-search algorithm for large-scale nonlinear programming},
  author={W{\"a}chter, Andreas and Biegler, Lorenz T},
  journal={Mathematical programming},
  volume={106},
  pages={25--57},
  year={2006},
  publisher={Springer}
}

@inproceedings{Agrawal2017,
    author = {Agrawal, Ayush and Sreenath, Koushil},
    title = {Discrete Control Barrier Functions for Safety-Critical Control of Discrete Systems with Application to Bipedal Robot Navigation},
    booktitle = {Proceedings of Robotics: Science and Systems},
    year = 2017
}

@ARTICLE{Li2024,
  author={Li, Yulin and Tang, Xindong and Chen, Kai and Zheng, Chunxin and Liu, Haichao and Ma, Jun},
  journal={IEEE Robotics and Automation Letters}, 
  title={Geometry-Aware Safety-Critical Local Reactive Controller for Robot Navigation in Unknown and Cluttered Environments}, 
  year={2024},
  volume={9},
  number={4},
  pages={3419-3426},
  doi={10.1109/LRA.2024.3360809}}

@ARTICLE{DHOCBF,
  author={Xiong, Yuhan and Zhai, Di-Hua and Tavakoli, Mahdi and Xia, Yuanqing},
  journal={IEEE Transactions on Cybernetics}, 
  title={Discrete-Time Control Barrier Function: High-Order Case and Adaptive Case}, 
  year={2023},
  volume={53},
  number={5},
  pages={3231-3239},
  doi={10.1109/TCYB.2022.3170607}}

\end{document}